\pdfoutput=1

\documentclass[11pt]{article}

\usepackage[final]{acl}

\usepackage{times}
\usepackage{latexsym}

\usepackage[T1]{fontenc}

\usepackage[utf8]{inputenc}

\usepackage{microtype}

\usepackage{inconsolata}

\usepackage{graphicx}

\usepackage{times}
\usepackage{latexsym}
\usepackage{amsmath} 
\usepackage{stfloats}
\usepackage{algorithm}
\usepackage{algorithmic}
\usepackage{multirow}
\usepackage{makecell}
\usepackage{arydshln}
\usepackage{amssymb} 
\usepackage{graphicx}
\usepackage{xcolor}
\usepackage{float}  
\usepackage{ulem}
\usepackage{hyperref}

\definecolor{darkred}{rgb}{0.6, 0, 0}  
\definecolor{darkblue}{rgb}{0.0, 0.0, 0.55}  
\definecolor{darkgreen}{RGB}{0,100,0} 

\title{Self-adaptive Multimodal Retrieval-Augmented Generation}

\author{Wenjia Zhai \\
  Unaffiliated \\
  \texttt{zhaiwenjia0311@163.com} \\}

\begin{document}
\maketitle
\begin{abstract}
Traditional Retrieval-Augmented Generation (RAG) methods are limited by their reliance on a fixed number of retrieved documents, often resulting in incomplete or noisy information that undermines task performance. Although recent adaptive approaches alleviated these problems, their application in intricate and real-world multimodal tasks remains limited. To address these, we propose a new approach called Self-adaptive Multimodal Retrieval-Augmented Generation (SAM-RAG), tailored specifically for multimodal contexts. SAM-RAG not only dynamically filters relevant documents based on the input query, including image captions when needed, but also verifies the quality of both the retrieved documents and the output. Extensive experimental results show that SAM-RAG surpasses existing state-of-the-art methods in both retrieval accuracy and response generation. By further ablation experiments and effectiveness analysis, SAM-RAG maintains high recall quality while improving overall task performance in multimodal RAG task. Our codes are available at \hyperlink{https://github.com/SAM-RAG/SAM_RAG}{https://github.com/SAM-RAG/SAM\_RAG}.

\end{abstract}

\section{Introduction}

Recent advances in large language models (LLMs) have significantly improved various natural language processing tasks \citep{manikandan2023language, openai2024gpt4technicalreport, ouyang2022training, touvron2023llama, anil2023palm2technicalreport}, including question answering \citep{tan2023can}. In addition, LLMs have begun to cross modal boundaries, exhibiting potential in various multimodal tasks such as visual comprehension \citep{pan2022amam} and code generation \citep{wang2024coderagbenchretrievalaugmentcode}. 

\begin{figure}[!t]  
    \centering
    \includegraphics[width=1\linewidth]{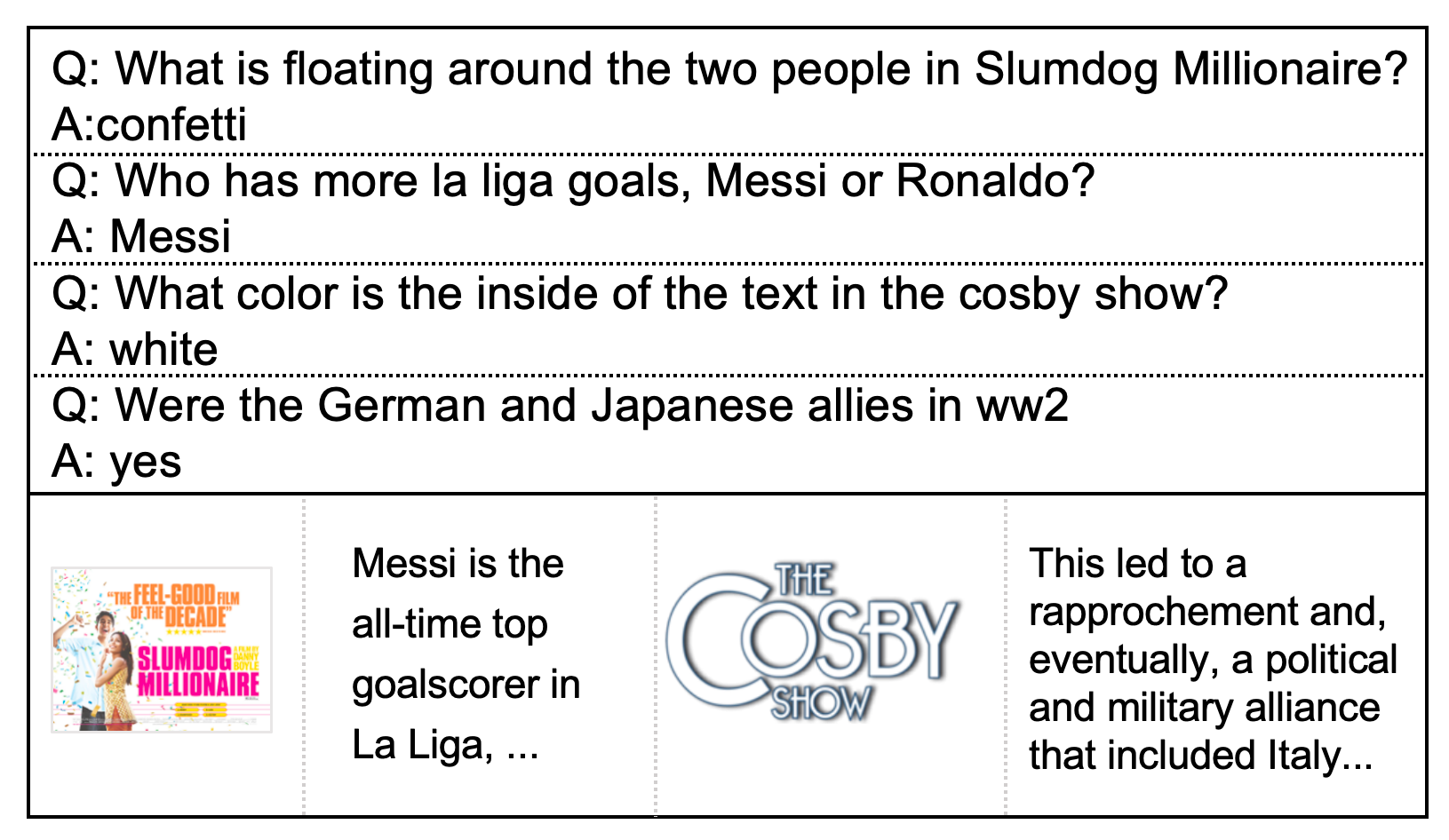}  
    \caption{
    \small  
    Some questions rely on visual information, which make text-only retrieval unfeasible. These questions require retrieving and reasoning over visual context.
    }
    \label{fig:overview}
\end{figure}

However, challenges remain, including the generation of false information (hallucination) \citep{li2022surveyretrievalaugmentedtextgeneration} and difficulties in knowledge updating \citep{zhang-etal-2023-large}, which hinder the applicability of a wider variety. To address these challenges, Retrieval-Augmented Generation (RAG) \citep{NEURIPS2020_6b493230} has been proposed, positing that documents semantically similar to a query likely contain the information needed to answer that query. By leveraging the in-context learning capabilities of LLMs, RAG enhances the likelihood of accurate responses from semantically related document retrievals while effectively reducing hallucinations and facilitating knowledge acquisition \citep{Li2024EnhancingLF}. As RAG technology evolves, it gradually expands into the multimodal domain, achieving progress in tasks involving texts \citep{wu2024retrieval}, images \citep{pan2022amam}, etc., an example of multimodal RAG task is shown in Figure \ref{fig:overview}.

RAG, while advantageous, encounters substantial obstacles, especially regarding adaptability \citep{zhao2024retrievalaugmentedgenerationaigeneratedcontent}. RAG's reliance on a fixed number of retrieved documents often results in retrieval omissions or excessive noise, negatively impacting performance, while it also lacks a mechanism to verify its generated responses. To address these challenges, adaptive RAG frameworks have been introduced, enabling dynamic adjustments based on task-specific needs, which greatly improve RAG performance in various tasks. However, research on adaptive RAG in multimodal contexts remains limited. 

To this end, we propose \textbf{S}elf-\textbf{A}daptive \textbf{M}ultimodal \textbf{R}etrieval-\textbf{A}ugmented \textbf{G}eneration (\textbf{SAM-RAG}), the first adaptive multimodal RAG framework capable of proactively selecting relevant data and self-evaluating generated responses. By employing knowledge distillation from state-of-the-art models to a smaller multimodal LLM, we ensure superior performance in multimodal tasks. SAM-RAG defines three essential criteria: relevance, usefulness, and support. In the course of processing queries, SAM-RAG assesses each set of retrieval outcomes separately, identifying pertinent data after retrieval, then generates a preliminary response based on these relevant contexts and assesses their contribution to response generation. Ultimately, the framework conducts a thorough evaluation of the query, contexts, and response. This process ensures that the generated answer is thoroughly substantiated by the provided contexts, thereby significantly reducing the occurrence of hallucinations.

We set up a series of experiments and introduce various baselines to evaluate our method. The experimental results indicate that SAM-RAG significantly outperforms the previous state-of-the-art framework like MuRAG \citep{chen2022murag} and other baselines in multimodal RAG tasks, demonstrating its robust capabilities in multimodal retrieval and generation. The ablation study evaluates the rationale and effectiveness of the three verifications, while the analysis of retrieval effectiveness underscores the importance of dynamic retrieval in enhancing performance. Additionally, the case study offers concrete examples demonstrating how each verification individually and collectively improves retrieval accuracy and the quality of answer generation.

The SAM-RAG framework integrates dynamic retrieval, relevance verification, and multi-stage answer validation to optimize multimodal tasks, ensuring accurate and supported answers while minimizing hallucinations. The process involves modality alignment, document retrieval, and iterative verifications for relevance, usability, and support, enhancing both output quality and model robustness. In summary, the contributions of this paper are as follows: 

\begin{enumerate}
    \item SAM-RAG introduces a dynamic multimodal RAG framework, enhancing relevant document selection while minimizing unnecessary retrievals; 
    \item It uses three key verification criteria — relevance, usability, and support — to improve retrieval accuracy and answer quality;
    \item Experiments show SAM-RAG's superior performance over state-of-the-art models, especially in multimodal tasks.
\end{enumerate}

\section{Related Works} 
\subsection{Retrieval-Augmented Generation}

RAG improved the relevance and accuracy of generated responses by incorporating external knowledge into language models. Early research focused on the text modality. Re2G \citep{glass-etal-2022-re2g} combined BM25 with deep learning retrieval, UPRISE \citep{cheng-etal-2023-uprise} introduced a prompt retriever to enhance the zero-shot capabilities of large language models, and QOQA \citep{koo2024optimizing} refined the retrieval performance through query regeneration. However, these methods relied on a fixed number of retrieval results, limiting adaptability to specific tasks. SAM-RAG overcomes this limitation by using adaptive retrieval, dynamically assessing recalled documents to improve response generation.

\subsection{Multimodal Retrieval-Augmented Generation} \label{sec:mrag}

A key area of RAG research is multimodal contexts, integrating various formats such as texts \citep{wu2024retrieval}, images \citep{pan2022amam}, tables \citep{dong2024ttc}, and audio \citep{xu2019learning}, with Visual Question Answering (VQA) being a significant focus \citep{ishmam2024image}. The main distinction between multimodal and textual RAG tasks is the necessity to unify diverse modalities into a single representation, known as modality alignment. Current LLMs exhibit superior reasoning, analysis, and generation capabilities by leveraging large-scale internet corpora, often outperforming visual and other modalities in these tasks \citep{jin2024efficientmultimodallargelanguage}. Thus, in multimodal RAG, a common approach involves converting non-textual modalities into text representations using a modality converter, thereby utilizing LLM's strengths in text processing \citep{zhao-etal-2023-retrieving}. Frameworks such as RA-VQA \citep{lin-byrne-2022-retrieval} used object detection to create multiple textual representations from images for standard RAG operations, while other approaches like MMHQA-ICL \citep{liu2023mmhqaiclmultimodalincontextlearning} and UniMMQA \citep{luo-etal-2023-unifying} combined text, tabular, and visual data to align with text-based RAG techniques. This paper specifically focuses on texts and images, incorporating these methods to extract image information effectively.

\subsection{Adaptive Retrieval-Augmented Generation}

The quality of retrieval significantly impacts RAG performance, as irrelevant or incorrect documents often cause "hallucinations" \citep{huang2023survey}. The conventional RAG relies on a fixed number of documents, which can miss crucial information or introduce irrelevant text. The conventional RAG also lacks verification mechanisms for generated answers. Adaptive RAG frameworks, such as Self-Improve \citep{huang-etal-2023-large} and Self-Refine \citep{NEURIPS2023_91edff07}, used feedback loops for error correction. Self-Correction \citep{welleck2023generating} and Self-Reasoning \citep{xia2024improvingretrievalaugmentedlanguage} further enhanced retrieval quality via reasoning. Self-RAG \citep{asai2024selfrag} improved adaptability by using on-demand retrieval and reflection. SAM-RAG builds on these ideas, introducing active screening and reflection mechanisms in the multimodal domain, creating a more dynamic retrieval process and advancing multimodal RAG applications.

\begin{figure*}[h!]
    \centering
    \includegraphics[width=1\linewidth]{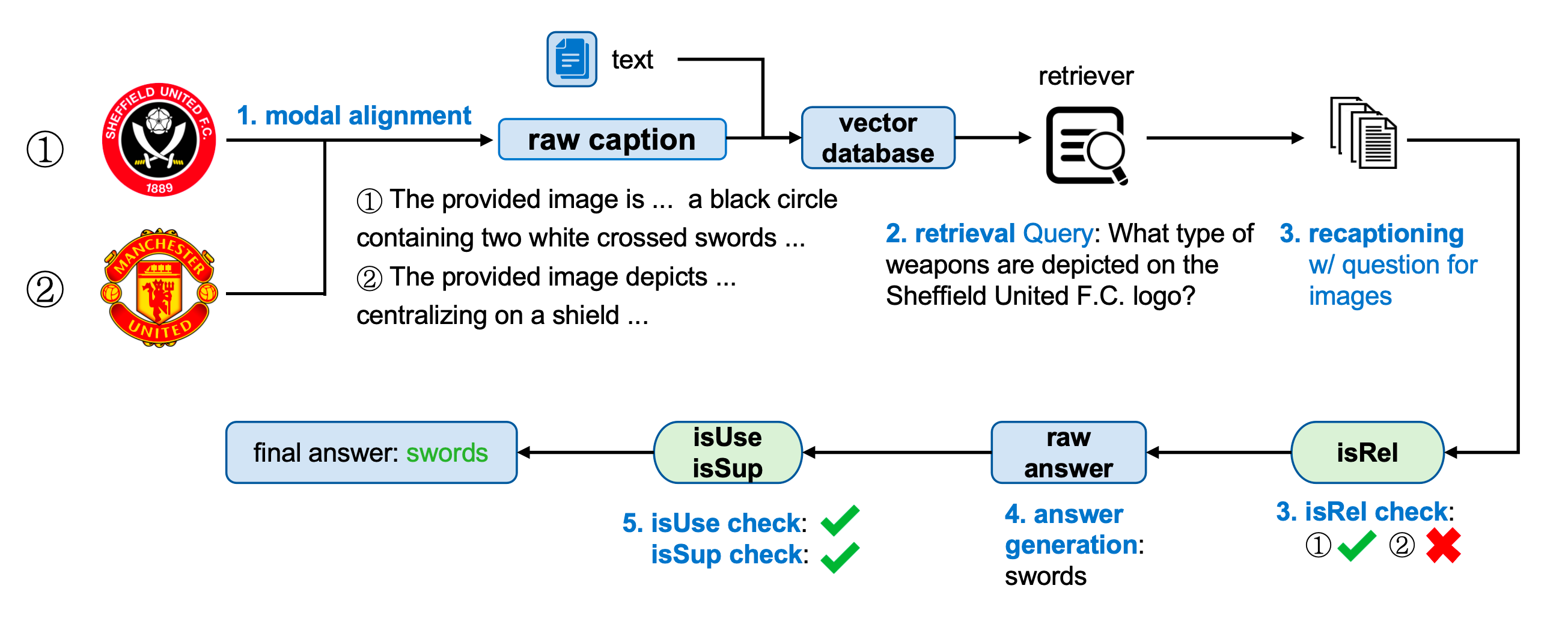}
    \caption{The illustration of SAM-RAG pipeline. If the documents are visual, they will be converted to textual captions, vectorized and stored with textual documents. When a retrieval is performed, documents are processed in batch based on semantic similarity. If a caption is in a batch, its coressponding image will be recaptioned with the query.}
    \label{fig:pipeline}
\end{figure*}

\section{Method}

In this study, we propose a novel multimodal RAG framework, SAM-RAG, as illustrated in Figure \ref{fig:pipeline}. SAM-RAG is designed to actively select multimodal data pertinent to the query for answer generation and to validate the generated responses from multiple perspectives, thereby ensuring the quality and reliability of the output.

\subsection{Task Definition}

Formally, the RAG task involves constructing a retrieval database from a document collection $\boldsymbol{D}=\{\boldsymbol{d}_1, \boldsymbol{d}_2, \ldots, \boldsymbol{d}_n\}$ using a retriever $\mathcal{R}$. For a set of queries $\boldsymbol{Q}=\{\boldsymbol{q}_1, \ldots, \boldsymbol{q}_m\}$, the retriever identifies a relevant document set $\boldsymbol{D}_{rel}=\{\boldsymbol{d}_1, \ldots, \boldsymbol{d}_k\}$ for each query $\boldsymbol{q}$, generating a sequence of $s$ tokens $\boldsymbol{A}=[\boldsymbol{a}_1, \ldots, \boldsymbol{a}_s]$ as the answer. Here, $k$ ($k < n$) is typically a predefined number for retrieval.

The multimodal RAG task typically involves the following steps: First, a modality converter $\mathcal{P}$ transforms information in other modalities into a format that the language model $\mathcal{L}$ can interpret, integrating it with textual data for retrieval. Next, for each query $\boldsymbol{q}$, the retriever $\mathcal{R}$ identifies the relevant information $\boldsymbol{D}_{rel}$. Finally, the language model $\mathcal{L}$ generates a sequence of $s$ tokens $[\boldsymbol{a}_1, \ldots, \boldsymbol{a}_s]$ as a response to the query $\boldsymbol{q}$, based on the retrieved information. The modules $\mathcal{P}$ and $\mathcal{L}$ are often combined into a single module $\mathcal{M}$ to streamline the entire process.

\subsection{Embedding Model}

In the RAG framework, the quality of retrieval results is critical to overall performance \citep{10.1145/3626772.3657957}. To enhance the recall rate of $\mathcal{R}$, we optimize its representational capabilities through contrastive learning.

For a given query $\boldsymbol{q}$, the supporting documents for the official answers constitute the positive sample set $\boldsymbol{D}_{pos}$, resulting in positive query-document pairs $(\boldsymbol{Q}, \boldsymbol{D}_{pos})$. The remaining documents form the negative sample set $\boldsymbol{D}_{neg}$. We employ dense passage retrieval (DPR) \citep{karpukhin-etal-2020-dense} to rank the negative document collection $\boldsymbol{D}_{neg}$ and randomly select 10 samples from the top 50 samples, thereby creating negative query-document pairs $(\boldsymbol{Q}, \boldsymbol{D}_{neg})$. This process culminates in a dataset of approximately 240k entries, with each query paired with corresponding a positive document $\boldsymbol{d}_{pos}$ and a negative document $\boldsymbol{d}_{neg}$, used to finetune the model $\mathcal{R}$.

We optimize the performance of $\mathcal{R}$ using the InfoNCE loss function \citep{he2020momentum}, as shown in Equation \ref{eq:infoNCE}, where $\boldsymbol{D}_{pos}$ represents the supporting documents for the answers, and $\tau$ is the temperature hyperparameter. The goal is to minimize $\ell$ between positive and negative samples.

\begin{equation}
  \label{eq:infoNCE}
  \ell = -\log \frac{\exp\left(\boldsymbol{Q} \cdot \boldsymbol{D}_{pos} / \tau \right)}{\sum \exp\left(\boldsymbol{Q} \cdot \boldsymbol{D} / \tau \right)}
\end{equation}

\subsection{Multimodal Knowledge Distillation}

Knowledge distillation transfers knowledge from a high-performance model (teacher) to a simpler model (student), enhancing the student's performance while reducing complexity \citep{yang2024survey}. In LLMs, this technique effectively passes capabilities from larger models to smaller ones \citep{jin2024efficientmultimodallargelanguage, gu2024minillm, li2023prompt}. We use \textbf{GPT} $\mathcal{G}$ to generate instruction-tuning data that improves our local model $\mathcal{M}$ \citep{peng2023instructiontuninggpt4}, refining its reasoning and generation skills. By leveraging chain-of-thought reasoning \citep{wei2022chain}, we retain inference processes of $\mathcal{G}$ for effective model distillation.

The data preparation for model distillation proceeds as follows. First, query-specific captions $\boldsymbol{D}_{sc}$ are generated by $\mathcal{G}$ for query-image pairs $\{\boldsymbol{q}, \boldsymbol{d}_{pos}\}$, and only the reasoning processes where the generated answers $\boldsymbol{A}$ match the standard answers are retained. Next, ${\textbf{\textcolor{darkred}{\fcolorbox{darkred}{white}{isRel}}}}$ is inferred using query-document pairs $\{\boldsymbol{q}, \boldsymbol{d} \mid \boldsymbol{d} \in {\boldsymbol{D}_{sc}, \boldsymbol{D}_t}\}$, keeping reasoning processes with ${\textbf{\textcolor{darkred}{\fcolorbox{darkred}{white}{isRel}}}}\texttt{=True}$. To improve efficiency, the negative set $\boldsymbol{D}_{neg}$ is ranked using DPR, and documents equal in number to the positive samples are randomly selected and inferenced from the top 10 ranked by similarity. The reasoning processes where ${\textbf{\textcolor{darkred}{\fcolorbox{darkred}{white}{isRel}}}}\texttt{=False}$ are retained. Coarse answers $\boldsymbol{A}$ are then generated from the filtered $\boldsymbol{D}_{pos}$, retaining reasoning processes where the conclusions align with the standard answers.

After generating $\boldsymbol{A}$, $\textbf{\textcolor{darkblue}{\fcolorbox{darkblue}{white}{isUse}}}\texttt{=?}$ is inferred based on the set $\{ \boldsymbol{q}, \boldsymbol{d}, \boldsymbol{A} \mid \boldsymbol{d} \in \{\boldsymbol{D}_{sc}, \boldsymbol{D}_t\} \}$. Using the same method as for ${\textbf{\textcolor{darkred}{\fcolorbox{darkred}{white}{isRel}}}}$, positive and negative samples with consistent inference results are retained. Finally, $\textbf{\textcolor{darkgreen}{\fcolorbox{darkgreen}{white}{isSup}}}\texttt{=?}$ is inferred. Similar to ${\textbf{\textcolor{darkred}{\fcolorbox{darkred}{white}{isRel}}}}$, for negative samples, reasoning processes where $\textbf{\textcolor{darkgreen}{\fcolorbox{darkgreen}{white}{isSup}}}\texttt{=False}$ are kept. While for positive samples, those with inference results of either \texttt{True} or \texttt{Partial} are retained.

\subsection{SAM-RAG}

The components of the SAM-RAG framework, as illustrated in Figure \ref{fig:pipeline}, are described. The overall process of executing multimodal RAG tasks is divided into five steps: \textbf{modality alignment}, \textbf{document retrieval}, \textbf{relevance verification}, \textbf{answer generation}, and \textbf{answer verification}. Image information from the entire document corpus is first aligned with the text modality. Upon receiving a query, document retrieval is performed. The retrieved results are then ranked by similarity, and each is assessed for relevance to the query, with only the relevant results being retained. Based on the relevant documents, the framework generates $\boldsymbol{A}$. The validity of the generated answer is subsequently verified, and if confirmed, the answer is returned, concluding the retrieval process. The core innovation of this method lies in the introduction of three key verification mechanisms, as shown in Table \ref{table:verifications}. Prompts are listed in Appendix \ref{app:prompt}.

\subsubsection{Modality Alignment}  
Raw captions $\boldsymbol{d}_{rc}$ are first generated using \textbf{GPT}, referred to as $\mathcal{G}$, without being tailored to specific queries. These captions are produced under constraints based on image titles. Since the raw captions are not specific to the queries, $\boldsymbol{d}_{rc}$ may omit key information relevant to the queries. Therefore, $\boldsymbol{d}_{rc}$ is used exclusively for indexing and similarity calculations. The generated $\boldsymbol{D}_{rc}$, along with the original textual documents $\boldsymbol{D}_t$, is then input into the embedding model $\mathcal{R}$ for vectorization, preparing for query retrieval.

\subsubsection{Document Retrieval} 
For a given query $\boldsymbol{q}$, scores for the document corpus $\boldsymbol{D}$ are obtained using DPR \citep{karpukhin-etal-2020-dense}, and the results are sorted in descending order of similarity scores for the next step.  

\subsubsection{Relevance Verification}

To track the retrieval status, a flag $\mathcal{F}$ is initialized to $\text{False}$, and a storage space $\boldsymbol{C}$ is created to hold relevant information. The retrieved documents are processed in batches, where the relevance of each retrieved text document $\boldsymbol{d}_t$ to the query is directly assessed using the relevance verification $\textbf{\textcolor{darkred}{\fcolorbox{darkred}{white}{isRel}}}$ (refer to Figure \ref{fig:pipeline} and Table \ref{table:verifications}, top row). If an image $\boldsymbol{d}_i$ is retrieved instead, a question-specific caption $\boldsymbol{d}_{sc}$ is generated based on the query, and its relevance is evaluated accordingly. When \textbf{\textcolor{darkred}{\fcolorbox{darkred}{white}{isRel}}}\texttt{=True}, the relevant information is stored in $\boldsymbol{C}$, and $\mathcal{F}$ is updated to $\text{True}$. After processing a batch, if \textbf{\textcolor{darkred}{\fcolorbox{darkred}{white}{isRel}}}\texttt{=False}, the next batch is processed; otherwise, the retrieval process is paused, and the next stage is initiated.

\subsubsection{Answer Generation}
The context $\boldsymbol{C}$ generated in the previous stage, along with the query $\boldsymbol{q}$, is utilized by the model $\mathcal{M}$ to produce a coarse answer $\boldsymbol{A}$ through an autoregressive generation method. Subsequently, the generated answer $\boldsymbol{A}$ undergoes a verification process to evaluate its relevance and support.

\begin{table*}[htbp!] 
  \centering
  \resizebox{0.8\linewidth}{!}{
  \begin{tabular}{lllll}
    \hline
    \textbf{Verification}         & \textbf{Definition}            & \textbf{Input} & \textbf{Output}      &\textbf{Action} \\  
    \hline
    \textbf{\textcolor{darkred}{\fcolorbox{darkred}{white}{isRel}}}        & If $\boldsymbol{d}$ is related to $\boldsymbol{q}$ & $\boldsymbol{q}$, $\boldsymbol{d}$           & \makecell[l]{\textbf{TRUE} \\ \textbf{FALSE}}                     &  \makecell[l]{$\mathcal{F}\gets \text{TRUE}, \boldsymbol{C} \leftarrow \boldsymbol{C} \cup \{d\}
$ \\ continueed}\\\hdashline
    \textbf{\textcolor{darkblue}{\fcolorbox{darkblue}{white}{isUse}}}      & If $\boldsymbol{A}$ accurately responded query $\boldsymbol{q}$ & $\boldsymbol{q}$, $\boldsymbol{A}$          & \makecell[l]{\textbf{TRUE} \\ \textbf{FALSE}}                       & \makecell[l]{proceeded to \textbf{\textcolor{darkgreen}{\fcolorbox{darkgreen}{white}{isSup}}}\\regenerated $\boldsymbol{A}$} \\\hdashline  
    \textbf{\textcolor{darkgreen}{\fcolorbox{darkgreen}{white}{isSup}}}         & If $\boldsymbol{A}$ supported by $\boldsymbol{C}$ & $\boldsymbol{q}$, $\boldsymbol{A}$, $\boldsymbol{C}$       & \makecell[l]{\textbf{TRUE} \\ \textbf{Partial}\\\textbf{FALSE}}    & \makecell[l]{returned $\boldsymbol{A}$\\ $\mathcal{F}\gets \text{FALSE}$ \\ $\mathcal{F}\gets \text{FALSE}$, clear $\boldsymbol{C}$} \\
    \hline
  \end{tabular}
  }
  \caption{\label{table:verifications}
    Explanation of each verification, its definition, inputs, outputs and corresponding actions. 
  }
\end{table*}

\subsubsection{Answer Verification}

In this stage, \textbf{\textcolor{darkblue}{\fcolorbox{darkblue}{white}{isUse}}} verification process (refer to Figure \ref{fig:pipeline} and Table \ref{table:verifications}, second row), is introduced to determine whether the generated answer $\boldsymbol{A}$ effectively addresses the query. If \textbf{\textcolor{darkblue}{\fcolorbox{darkblue}{white}{isUse}}}\texttt{=True}, the next step will involve evaluating the support of $\boldsymbol{A}$ using \textbf{\textcolor{darkgreen}{\fcolorbox{darkgreen}{white}{isSup}}} (refer to Figure \ref{fig:pipeline} and Table \ref{table:verifications}, third row); otherwise, $\boldsymbol{A}$ is regenerated based on $\boldsymbol{C}$. The purpose of \textbf{\textcolor{darkgreen}{\fcolorbox{darkgreen}{white}{isSup}}} is to confirm that $\boldsymbol{A}$ is adequately supported by $\boldsymbol{C}$, thereby preventing situations where $\boldsymbol{A}$ lacks support from $\boldsymbol{D}_{rel}$. If \textbf{\textcolor{darkgreen}{\fcolorbox{darkgreen}{white}{isSup}}}\texttt{=True}, it is indicated that $\boldsymbol{A}$ fulfills the task requirements, and $\boldsymbol{A}$ is returned, concluding the retrieval process. If \textbf{\textcolor{darkgreen}{\fcolorbox{darkgreen}{white}{isSup}}}\texttt{=False}, it signifies that $\boldsymbol{A}$ is unsupported by $\boldsymbol{C}$, prompting a reset of both $\boldsymbol{C}$ and $\mathcal{F}$ to return to the first stage for further retrieval. If \textbf{\textcolor{darkgreen}{\fcolorbox{darkgreen}{white}{isSup}}}\texttt{=Partial}, it suggests incomplete support information, leading to the retention of $\boldsymbol{C}$ while resetting $\mathcal{F}$, and returning to the first stage to continue the retrieval. The objective of this stage is to ensure that the final output aligns with task requirements and minimizes the risk of potential hallucinations.

Through the aforementioned design, the SAM-RAG method ensures effective retrieval while minimizing the risk of generating misleading answers, thereby safeguarding the quality of the final output. To enhance the model's stability and the reliability of the results, we incorporate the Chain-of-Thought method \citep{wei2022chain} into each verification process and the answer generation step. Furthermore, we can implement the self-consistency strategy \citep{wang2023selfconsistency} to further bolster robustness.

\section{Experiments}

\subsection{Dataset}

In this study, the MultimodalQA dataset \citep{talmormultimodalqa} is used as a benchmark to evaluate the performance of the SAM-RAG framework. This dataset comprises multimodal question-answering inputs, including texts, images, and tables. To facilitate a fair comparison with the leading multimodal RAG model, MuRAG \citep{chen2022murag}, the experiments focus exclusively on the TextQ and ImageQ categories of questions within the MultimodalQA dataset.

\subsection{Metrics}

F1 score and exact match (\textbf{EM}) metrics \citep{rajpurkar2016squad} are employed to evaluate the quality of the generated answers. Additionally, two metrics are used to evaluate retrieval performance: Recall@N (as defined in Equation \ref{eq:recall@n}) and a newly introduced metric, the average retrieval number (\textbf{ARN}), which measures the average number of documents retrieved to generate the final answer, providing a more nuanced assessment of retrieval efficiency in SAM-RAG.

\begin{equation}
  \label{eq:recall@n}
  \text{recall@N} = 
\begin{cases}
1, & \text{if } \boldsymbol{D}_{pos} \subseteq  \mathcal{R}_{q_i}(\boldsymbol{D}) \\
0, & \text{otherwise}
\end{cases}
\end{equation}

\subsection{Experimental Settings}

\begin{table*}[htbp]
\renewcommand{\arraystretch}{1.1}
  \centering
  \begin{tabular}{l}
    \hline
    \textbf{Evaluation} \\
    \textbf{Metrics}    \\
    \hline
    \textbf{MuRAG}             \\\hline
    \textbf{RAG($\mathcal{R}$ + $\mathcal{M}$)}         \\
    \textbf{RAG($\mathcal{R}^*$ + $\mathcal{M}$)}       \\
    \textbf{RAG($\mathcal{R}$ + $\mathcal{M}^*$)}               \\
    \textbf{RAG($\mathcal{R}^*$ + $\mathcal{M}^*$)}                    \\\hline
    \textbf{SAM-RAG($\mathcal{R}^*$ + $\mathcal{M}$)}           \\
    \textbf{SAM-RAG($\mathcal{R}^*$ + $\mathcal{M}^*$)}                     \\\hline
    \textbf{RAG($\mathcal{R}$ + $\mathcal{G}$)}             \\
    \textbf{RAG($\mathcal{R}^*$ + $\mathcal{G}$)}                  \\
    \textbf{SAM-RAG($\mathcal{R}^*$ + $\mathcal{G}$)}               \\\hline
  \end{tabular}
  \begin{tabular}{cc}
    \hline
    \multicolumn{2}{c}{\textbf{Text}} \\
    \textbf{F1}   & \textbf{EM} \\
    \hline
    56.10             & 49.70      \\\hline
    35.47             & 32.44      \\
    40.82             & 38.85      \\
    42.86             & 41.66          \\
    52.31             & 51.18          \\\hline
    40.49             & 39.25          \\
    54.35             & 52.74          \\\hline
    60.92             & 58.88          \\
    66.64             & 65.49          \\
    \textbf{71.03}             & \textbf{70.10}          \\\hline
  \end{tabular}
  \begin{tabular}{cc}
    \hline
    \multicolumn{2}{c}{\textbf{Image}} \\
    \textbf{F1} & \textbf{EM}  \\\hline
    56.50         & 56.50      \\\hline
    33.56         & 30.95      \\
    36.20         & 33.81      \\
    39.26             & 37.62          \\
    46.20             & 44.29          \\\hline
    39.54             & 38.57          \\
    54.13             & 52.38          \\\hline
    54.11             & 50.47           \\
    59.09             & 57.62           \\
    \textbf{80.51}             & \textbf{79.98}           \\\hline
  \end{tabular}
  \begin{tabular}{c}
    \hline
    \textbf{All}      \\
    \textbf{EM}       \\\hline
    51.40             \\\hline
    32.12             \\
    37.75             \\
    40.77             \\
    47.97             \\\hline
    39.10             \\
    52.66             \\\hline
    57.04             \\
    63.92             \\
    \textbf{72.26}    \\\hline
  \end{tabular}
  \caption{Multimodal test-set synthesis evaluation results. "RAG" means conventional RAG with retrieval top k as 8. $\mathcal{R}$ and $\mathcal{M}$ indicate model is not finetuned; $\mathcal{R}^*$ and $\mathcal{M}^*$ indicate model is finetuned; $\mathcal{G}$ indicates model is \textbf{GPT}.}
  \label{tab:syn_eval}
\end{table*}

The experimental setup involves the finetuning of both the embedding model $\mathcal{R}$ and the vision-language model $\mathcal{M}$, alongside generation and inference tasks using either $\mathcal{M}$ or $\mathcal{G}$. Finetuning is performed on a single NVIDIA A100 80GB GPU, with the server running Ubuntu Server 20.04 LTS.

For the embedding model, \texttt{bge-base-en-v1.5}\footnote{https://huggingface.co/BAAI/bge-base-en-v1.5} \citep{xiao2024c} is used, finetuned using the FlagEmbedding framework\footnote{https://github.com/FlagOpen/FlagEmbedding} \citep{bge_embedding}. Key hyperparameters include a learning rate of 1e-5, 5 training epochs, and a batch size of 64. The vision-language model $\mathcal{M}$, \texttt{LLaVA-v1.5-7b}\footnote{https://huggingface.co/llava-hf/llava-1.5-7b-hf} \citep{liu2024visual}, is finetuned via the LLaMA-Factory framework\footnote{https://github.com/hiyouga/LLaMA-Factory} \citep{zheng2024llamafactory} using LoRA \citep{hu2022lora}. The setup includes a learning rate of 1e-4, 5 epochs, a batch size of 1, and 8 gradient accumulation steps, with AdamW \citep{loshchilov2017decoupled} as the optimizer. 

During data generation and inference, $\mathcal{G}$\footnote{\textbf{GPT} version: GPT-4o-2024-05-13} is utilized, with a temperature parameter set to 1.2 to ensure response diversity, and is further used to finetune $\mathcal{M}$ and conduct inference tasks.

\subsection{Main Results}
The experimental results, summarized in Table \ref{tab:syn_eval}, demonstrate that SAM-RAG achieves superior performance across all evaluated tasks, significantly outperforming baseline models, including the state-of-the-art MuRAG. Below, we provide a detailed analysis of the performance improvements and insights gained from the comparisons.

\begin{enumerate}
\item \textbf{Overall Performance}: Across both text and image tasks, SAM-RAG delivers higher F1 and EM scores compared to baseline methods. These improvements underscore the effectiveness of the dynamic retrieval and generation mechanisms of SAM-RAG. The roles of key components, \textbf{\textcolor{darkred}{\fcolorbox{darkred}{white}{isRel}}}, \textbf{\textcolor{darkblue}{\fcolorbox{darkblue}{white}{isUse}}} and \textbf{\textcolor{darkgreen}{\fcolorbox{darkgreen}{white}{isSup}}}, are further analyzed in the ablation study (\ref{ablation}) and illustrated through case studies (\ref{case}, Figure \ref{fig:case}). Additional case examples are provided in Appendix \ref{app:cases}.
\item \textbf{Performance Comparison with Baselines}: Table \ref{tab:syn_eval} (top) demonstrates that fine-tuning the retrieval model $\mathcal{R}$ or the VLM $\mathcal{M}$ improves multimodal RAG performance. However, fine-tuning $\mathcal{M}$ provides more substantial gains. This suggests that the baseline retrieval model $\mathcal{R}$ is already sufficiently optimized, leading to smaller performance improvements from fine-tuning. In contrast, optimizing $\mathcal{M}$ directly enhances the model’s ability to process multimodal data. Notably, the fine-tuned \textbf{RAG($\mathcal{R}^*$ + $\mathcal{M}^*$)} shows slightly lower performance than MuRAG, despite differences in the volume of training data.
\item \textbf{SAM-RAG vs. Conventional RAG}: As shown in Table \ref{tab:syn_eval} (middle), SAM-RAG consistently outperforms conventional RAG methods. By dynamically retrieving documents until relevant information is found, SAM-RAG avoids the limitations of fixed retrieval strategies. Fine-tuning the retrieval model $\mathcal{R}$ speeds up document retrieval but does not significantly impact overall performance, as the strength of SAM-RAG lies in leveraging the fine-tuned $\mathcal{M}$. The results indicate that fine-tuning $\mathcal{M}$ leads to the most notable performance improvements in SAM-RAG, especially for multimodal tasks.
\item \textbf{Effect of GPT Integration}: Table \ref{tab:syn_eval} (bottom) highlights the substantial performance gains from integration of $\mathcal{G}$ into the SAM-RAG and conventional RAG frameworks. Models incorporating $\mathcal{G}$ outperform all other configurations, underscoring $\mathcal{G}$'s advanced reasoning and understanding abilities. SAM-RAG combined with $\mathcal{G}$ shows the most significant improvements, particularly in visual tasks, where it exceeds textual performance. This suggests that the SAM-RAG framework, when paired with $\mathcal{G}$, achieves a deeper understanding of visual content than other approaches.
\end{enumerate}

\subsection{Ablation Study} \label{ablation}

\begin{table}[htbp!]
\renewcommand{\arraystretch}{1.2}
  \centering
  \resizebox{\linewidth}{!}{
    \begin{tabular}{l}
                                       \hline
      \textbf{Experiment}              \\\hline
      \textbf{MuRAG}                   \\\hline
      \textbf{RAG}                     \\\hline
      \textbf{W/ \textbf{\textcolor{darkred}{\fcolorbox{darkred}{white}{isRel}}}}              \\
      \textbf{WO/ \textbf{\textcolor{darkgreen}{\fcolorbox{darkgreen}{white}{isSup}}}}           \\
      \textbf{WO/ \textbf{\textcolor{darkblue}{\fcolorbox{darkblue}{white}{isUse}}}}            \\
      \textbf{W/ all}                \\\hline
    \end{tabular}
    \begin{tabular}{c}                 \hline
      \textbf{F1}                      \\\hline
      -                                \\\hline
      64.99                            \\\hline
      65.62                            \\
      68.56                            \\
      69.05                            \\
      \textbf{73.19}                   \\\hline
    \end{tabular}
    \begin{tabular}{c}                 \hline
         $\textcolor{darkred}{\Delta{\textbf{F1}}}$       \\\hline
         -                             \\\hline
         \textcolor{darkred}{-8.20}                        \\\hline
         \textcolor{darkred}{-7.57}                         \\
         \textcolor{darkred}{-4.63}                         \\
         \textcolor{darkred}{-4.14}                         \\
         \textcolor{darkred}{-}                         \\\hline
    \end{tabular}
    \begin{tabular}{c}                 \hline
      \textbf{EM}                      \\\hline
      51.40                            \\\hline
      63.92                            \\\hline
      64.96                            \\
      67.07                            \\
      67.99                            \\
      \textbf{72.26}                   \\\hline
    \end{tabular}
    \begin{tabular}{c}                 \hline
         $\textcolor{darkred}{\Delta{\textbf{EM}}}$       \\\hline
         \textcolor{darkred}{-20.86}                        \\\hline
         \textcolor{darkred}{-8.84}                         \\\hline
         \textcolor{darkred}{-7.30}                         \\
         \textcolor{darkred}{-5.69}                         \\
         \textcolor{darkred}{-4.27}                         \\
         \textcolor{darkred}{-}                         \\\hline
    \end{tabular}
    }
  \caption{
  Effect of different verification combinations.\\"RAG" indicates conventional RAG pipeline. \\\textcolor{darkred}{$\Delta$}: The difference is between the best value and the corresponding value.
}
  \label{tab:diff_judge}
\end{table}

To validate the effectiveness of each component of SAM-RAG, a series of ablation studies are performed that check the performance of the framework after removing each component, as shown in Table \ref{tab:diff_judge}. The results indicate that, compared to conventional RAG, the introduction of distinct verification, particularly \textbf{\textcolor{darkred}{\fcolorbox{darkred}{white}{isRel}}} and \textbf{\textcolor{darkgreen}{\fcolorbox{darkgreen}{white}{isSup}}}, leads to notable improvements in both F1 and EM scores. Specifically, although introducing \textbf{\textcolor{darkred}{\fcolorbox{darkred}{white}{isRel}}} only brings minimal improvement, the subsequent introduction of \textbf{\textcolor{darkblue}{\fcolorbox{darkblue}{white}{isUse}}} and \textbf{\textcolor{darkgreen}{\fcolorbox{darkgreen}{white}{isSup}}} makes a greater impact. When all verifications are combined (labeled "with all"), the EM score rises to the maximum value, reflecting an improvement of more than 20\% compared to the performance of MuRAG. This finding suggests that simultaneous consideration of relevance (\textbf{\textcolor{darkred}{\fcolorbox{darkred}{white}{isRel}}}), support (\textbf{\textcolor{darkgreen}{\fcolorbox{darkgreen}{white}{isSup}}}), and usability (\textbf{\textcolor{darkblue}{\fcolorbox{darkblue}{white}{isUse}}}) allows the model to effectively filter the most valuable information for the generation of answers, significantly improving the quality of the output.

The introduction of even a single verification leads to an increase in the model's EM score, demonstrating that each criterion contributes uniquely to optimization. However, the results indicate that the most substantial improvements occur when multiple verifications are employed in combination. This suggests that integrating various verification mechanisms can comprehensively optimize information retrieval and generation processes, thereby minimizing the impact of irrelevant or low-quality information.

\subsection{Analysis of Retrieval Effectiveness}

\begin{figure}[htbp]
    \centering
    \includegraphics[width=1\linewidth]{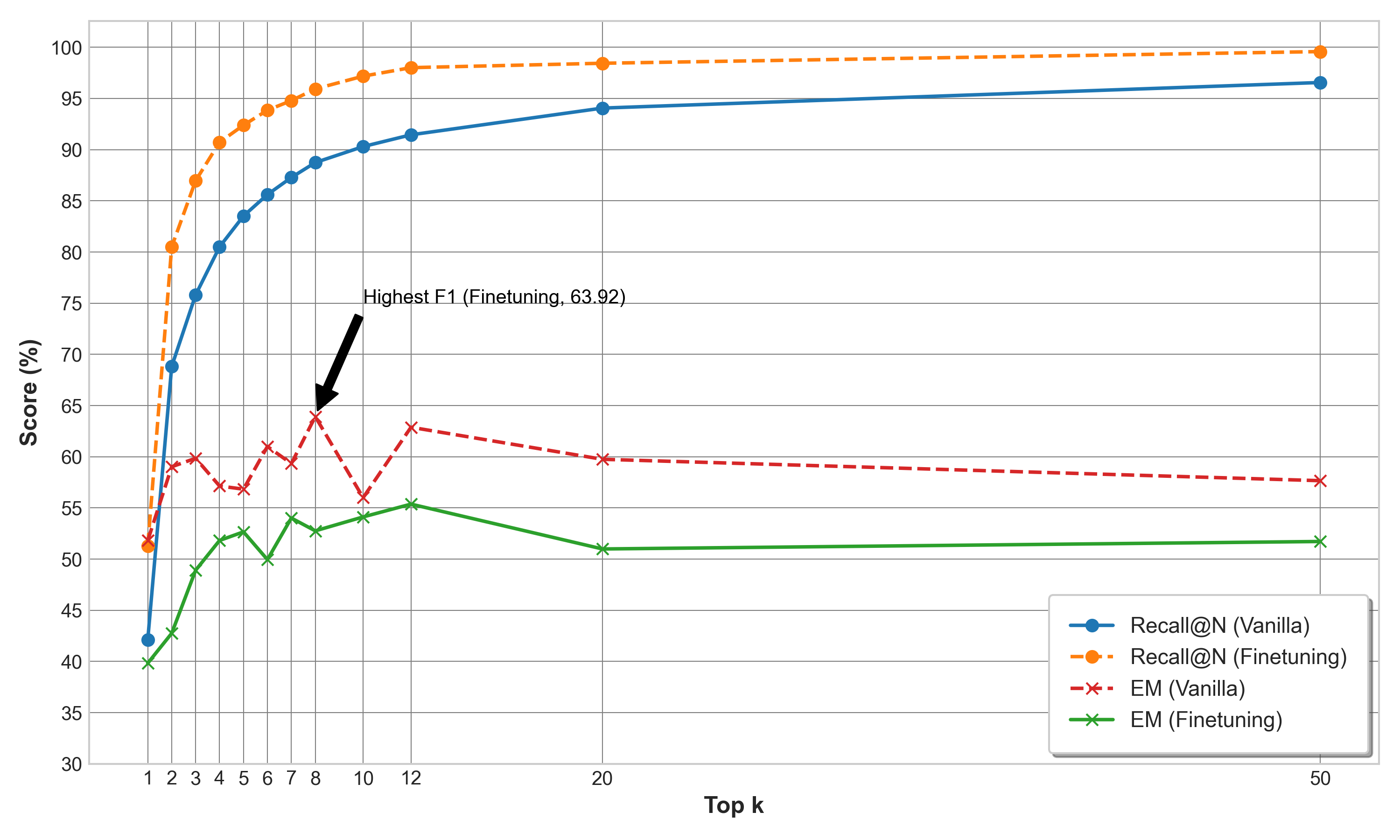}
    \caption{Effect of the Different Retrieval Numbers on Recall@N and EM.}
    \label{fig:retr_num}
\end{figure}

To assess retrieval effectiveness, we analyzed the top-performing SAM-RAG and conventional RAG models, as shown in Figure \ref{fig:retr_num} and Table \ref{tab:retr_eval}. The results indicate that $\mathcal{R}^*$ consistently outperforms $\mathcal{R}$ in Recall@N across all top $k$ values, suggesting that fine-tuning the embeddings improves relevant context retrieval. Table \ref{tab:syn_eval} further shows that the EM scores for $\mathcal{R}^*$ are generally higher than for $\mathcal{R}$. Notably, at a top $k$ of 8, the EM score surpasses all scores of $\mathcal{R}$ by nearly 20\%, suggesting that fine-tuning $\mathcal{R}$ benefits more from a moderate number of retrievals, while too many retrievals may introduce noise. Unlike conventional RAG, which often retrieves more documents than needed thus introduces irrelevant content, SAM-RAG's recall numbers closely match the "Gold Reference" values and yield higher EM scores for both text and image retrieval. This suggests that SAM-RAG effectively retrieves relevant information while reducing noise, resulting in more accurate and higher-quality outputs.

\begin{table}[htbp] 
  \centering
  \resizebox{\linewidth}{!}{
  \begin{tabular}{l}                     \hline
    \textbf{Evaluation}                  \\
    \textbf{Metrics}                     \\\hline
    \textbf{Gold Reference}              \\\hline
    \textbf{conventional RAG}                 \\
    \textbf{SAM-RAG}                         \\\hline
  \end{tabular}
  \begin{tabular}{cc}
    \hline
    \multicolumn{2}{c}{\textbf{Text}}    \\
    \textbf{EM}       & \textbf{ARN}     \\\hline
    -                 & 1.55             \\\hline
    65.69             & 7                \\
    \textbf{70.01}    & \textbf{1.77}    \\\hline
  \end{tabular}
  \begin{tabular}{cc}                    \hline
    \multicolumn{2}{c}{\textbf{Image}}   \\
    \textbf{EM}       & \textbf{ARN}     \\\hline
    -                 & 1.00             \\\hline
    57.62             & 7                \\
    \textbf{80.00}    & \textbf{1.24}    \\\hline
  \end{tabular}
  }
  \caption{Multimodal test-set synthesis evaluation results and recall@N analysis. \\"\textbf{Gold Reference}" means the texts and images which are labeled as "supporting context"; "\textbf{ARN}" means the average number of retrieved documents for the answers.}
  \label{tab:retr_eval}
\end{table}

\subsection{Case Study} \label{case}

\begin{figure*}[htbp!]
    \centering
    \includegraphics[width=1\linewidth]{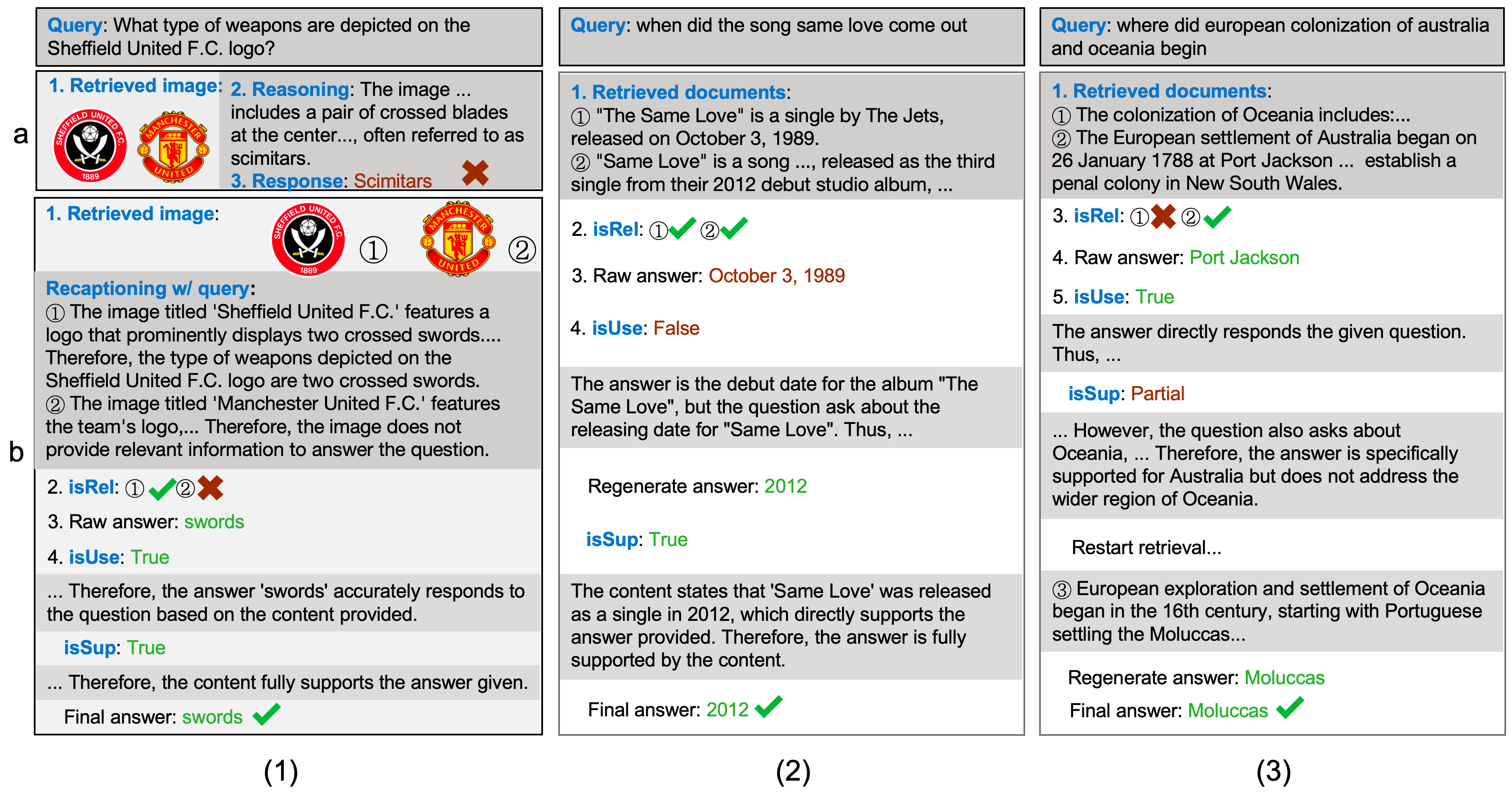}
    \caption{Complete real cases. (1) is a comparison between (a) convntional RAG and (b) SAM-RAG. (2) is a case where $\textbf{\textcolor{darkblue}{\fcolorbox{darkblue}{white}{isUse}}}$ is \texttt{False}. (3) is a case where $\textbf{\textcolor{darkgreen}{\fcolorbox{darkgreen}{white}{isSup}}}$ is \texttt{Partial}. For simplicity, only two visual documents are displayed.}
    \label{fig:case}
\end{figure*}

To intuitively illustrate the effectiveness of SAM-RAG, we present selected examples in Figure \ref{fig:case}. The left panel contrasts the standard RAG pipeline with SAM-RAG. In (a), the conventional RAG retriever $\mathcal{R}$ retrieves multiple documents, but despite including relevant content (e.g., the Sheffield United F.C. logo), $\mathcal{M}$ is misled by irrelevant information, resulting in incorrect generation. In (b), SAM-RAG mitigates this problem by filtering irrelevant documents through \textbf{\textcolor{darkred}{\fcolorbox{darkred}{white}{isRel}}}, ensuring that $\mathcal{M}$ focuses only on relevant information, leading to accurate generation based on essential content like $\boldsymbol{d}_{sc}$.

The middle panel highlights how $\mathcal{M}$ uses the \textbf{\textcolor{darkblue}{\fcolorbox{darkblue}{white}{isUse}}} to review its generated answer. Initially, $\mathcal{M}$ produces an incorrect response but recognizes the error and self-corrects, generating the correct answer.

In the right panel, after initial verifications, the \textbf{\textcolor{darkgreen}{\fcolorbox{darkgreen}{white}{isSup}}} identifies a gap between the content and the generated answer. This prompts $\mathcal{R}$ to retrieve additional information, enabling a more complete and accurate response.

This work presents SAM-RAG, a novel multimodal RAG framework that filters and analyzes retrieved content. SAM-RAG converts image documents into text, then assesses their relevance to the query. Once relevant documents are identified, it generates initial answers and evaluates how these documents contribute to the answer generation, determining if they fully support it. Experimental results show that SAM-RAG outperforms current state-of-the-art models on benchmark datasets.

\section{Limitations} 

The SAM-RAG framework, while innovative in its approach to multimodal retrieval-augmented generation, is not without limitations. 

One notable limitation is the potential for biased outputs, which can arise from the training data used for the underlying models. If the data contains historical biases or unrepresentative samples, the SAM-RAG system may inadvertently generate responses that unfair narratives, impacting its effectiveness and fairness.

Additionally, the reliance on large language models poses significant data privacy challenges. Given that these models are trained on extensive datasets, there is a risk that sensitive or personally identifiable information could be inadvertently incorporated into the responses. If not properly managed, this could lead to privacy breaches or the dissemination of confidential information, raising ethical concerns about data handling and user trust.

Moreover, the framework's complexity can also introduce safety issues. The multi-step processes involved in retrieval, relevance verification, and answer generation may not always guarantee that the final output aligns perfectly with task requirements, potentially leading to misleading or hallucinated information. This underscores the need for rigorous validation mechanisms to ensure the reliability of generated answers.

\bibliography{custom}

\newpage
\onecolumn
\appendix

\textbf{Appendix}

\setcounter{figure}{0}
\renewcommand{\thefigure}{A\arabic{figure}}

\newpage
\section{Typical Cases}
\label{app:cases}
Due to the black box nature of an LLM, here the typical cases are not analytic or explanatory.
\subsection{$\textbf{isRel}$}

\subsubsection{Positive}
\begin{figure}[htbp]  
    \centering
    \includegraphics[width=0.4\linewidth]{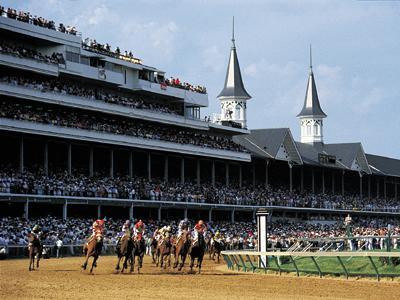}  
    \caption{
    \small  
    Image title: Kentucky Derby
    }
    \label{fig:isRel_pos}
\end{figure}

Retrieved image: (Figure. \ref{fig:isRel_pos})

\begin{quote}
    \textbf{Question}: What animals race in the Kentucky Derby? \\
    \textbf{$\textbf{\textcolor{darkred}{\fcolorbox{darkred}{white}{isRel}}}$ Response}: \{'Reasoning': 'The image of the Kentucky Derby prominently features horses racing on a track with a crowd watching from the stands. The Kentucky Derby is a well-known horse racing event, which is explicitly illustrated by the horses visible in the image. This visual directly connects to the question by showcasing the animals, horses, that race in the Kentucky Derby. Therefore, the image content answers the question satisfactorily.', 'Response': 'True'\} 
\end{quote}
$\texttt{found}\to{True}$
\\\textbf{Continue searching until the end of current batch...}
\subsubsection{Negative}
\begin{figure}[htbp]  
    \centering
    \includegraphics[width=0.2\linewidth]{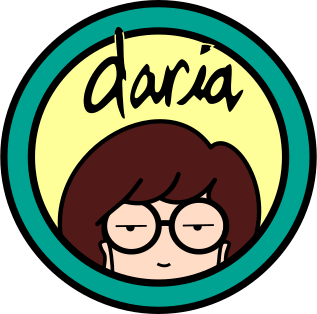}  
    \caption{
    \small  
    Image title: Daria
    }
    \label{fig:isRel_neg}
\end{figure}

Retrieved image: (Figure. \ref{fig:isRel_neg})

\begin{quote}
    \textbf{Question}: What object is in front of Felicia Day's face? \\
    \textbf{\textcolor{darkred}{\fcolorbox{darkred}{white}{isRel}}}\ \textbf{Response}: \{'Reasoning': "The image title is 'Daria,' which refers to a character from an animated series, and it visually features a character with glasses. The question asks about an object in front of Felicia Day's face which is unrelated to either the 'Daria' series or its content. The image does not depict Felicia Day, and there is no visible object in front of a character's face named Felicia Day here. Therefore, there is no relation between the image titled 'Daria' and the question about Felicia Day.", 'Response': 'False'\} 
\end{quote}
\textbf{Continue searching...}

\subsection{isUse}
\subsubsection{Positive}
\begin{quote}
    \textbf{Question}: What is the name of the final album released by the Christian rock band that formed in the mid 1980s at Kentucky Christian University in Grayson, Kentucky?  \\
    \textbf{Supporting Context}: Adios: The Greatest Hits is the final album released by Christian rock band Audio Adrenaline, and their second Greatest Hits album.\\
    \textbf{Draft answer}: Adios: The Greatest Hits \\
    \textbf{\textcolor{darkblue}{\fcolorbox{darkblue}{white}{isUse}}}\ \textbf{Response}: \{'Reasoning': "The answer 'Adios: The Greatest Hits' correctly identifies the name of the final album released by the Christian rock band Audio Adrenaline, as mentioned in the source content. The content specifies that 'Adios: The Greatest Hits' is the final album, aligning logically with the requirement to name the final album in the question. Additionally, the band, although not explicitly mentioned in the answer, is implicitly known from the context of the content and relevant to the question. Both the format and logical consistency criteria are met.", 'Response': 'True'\}
\end{quote}
\textbf{Proceed to isSup...}

\subsubsection{Negative}
\begin{quote}
    \textbf{Question}: are all of the songs in mamma mia here we go again by abba \\
    \textbf{Supporting Context}: Mamma Mia! Here We Go Again was announced on May 19, 2017, with a release date of July 20, 2018. It was written and directed by Ol Parker. On September 27, 2017, Benny Andersson confirmed 3 ABBA songs that would be featured in the film: "When I Kissed the Teacher," "I Wonder (Departure)," and "Angeleyes." "I Wonder (Departure)" was cut from the film, but is included on the soundtrack album.\\
    \textbf{Draft answer}: No \\
    \textbf{\textcolor{darkblue}{\fcolorbox{darkblue}{white}{isUse}}}\ \textbf{Response}: \{'Reasoning': "The answer 'No' implies that not all songs in 'Mamma Mia! Here We Go Again' are by ABBA, but the content does not explicitly affirm this, nor does it list songs by artists other than ABBA. Although it mentions some ABBA songs included in the film, it lacks specific information about non-ABBA songs. Thus, without specific supporting content to verify that not all songs are by ABBA, the answer cannot be logically consistent with the provided information.", 'Response': 'False'\}
\end{quote}
\textbf{Regenerate answer...}
\subsection{isSup}
\subsubsection{Postive}

\begin{quote}
    \textbf{Question}: what are flying monkeys in wizard of oz called  \\
    \textbf{Supporting Context}: Winged monkeys (often referred to in adaptations and popular culture as flying monkeys) are fictional characters created by American author L. Frank Baum in his classic children's novel The Wonderful Wizard of Oz (1900). The winged monkeys are exactly what the name implies: jungle monkeys with powerful bird-like feathered wings attached to their shoulders which allow them to fly. They are most notable from the famous 1939 musical movie by MGM. Ever since, they have taken their own place in popular culture, regularly referenced in comedic or ironic situations as a source of evil or fear.\\
    \textbf{Draft answer}: winged monkeys \\
    \textbf{\textcolor{darkgreen}{\fcolorbox{darkgreen}{white}{isSup}}}\ \textbf{Response}: \{'Reasoning': "The content specifies that the characters in 'The Wonderful Wizard of Oz' referred to as flying monkeys in adaptations and popular culture, are originally called 'winged monkeys' in the novel. Hence, the given answer aligns correctly with the content.", 'Response': 'True'\}
\end{quote}
\textbf{Return the answer}
\subsubsection{Partial}
\begin{quote}
    \textbf{Question}: who was first offered the part of columbo  \\
    \textbf{Supporting Context}: In 1968, the same play was made into a two-hour television movie that aired on NBC. The writers suggested Lee J. Cobb and Bing Crosby for the role of Columbo, but Cobb was unavailable and Crosby turned it down because he felt it would take too much time away from the golf links. Director Richard Irving convinced Levinson and Link that Falk, who wanted the role, could pull it off even though he was much younger than the writers had in mind.\\
    \textbf{Draft answer}: Lee J. Cobb \\
    \textbf{\textcolor{darkgreen}{\fcolorbox{darkgreen}{white}{isSup}}}\ \textbf{Response}: \{'Reasoning': "The content mentions that Lee J. Cobb was suggested by the writers for the role of Columbo, indicating that he was among the first choices for the role. However, it doesn’t clearly state he was formally offered the part, as Bert Freed was the first actor to portray Columbo. Since it's indicated Freed already played the role and Lee J. Cobb was just suggested later, the interpretation of 'offered' here could be nuanced. Given this ambiguity but based on the information that Cobb was suggested, the content supports that proposition within this context.", 'Response': 'Partial'\}
\end{quote}
\textbf{Reset draft answer and restart searching on next batch...}

\subsubsection{Negative}
\begin{quote}
    \textbf{Question}: who was considered the father of the blues  \\
    \textbf{Supporting Context}: McKinley Morganfield (April 4, 1913 April 30, 1983), known professionally as Muddy Waters, was an American blues singer-songwriter and musician who is often cited as the "father of modern Chicago blues", and an important figure on the post-war blues scene. His style of playing has been described as "raining down Delta beatitude".\\
    \textbf{Draft answer}: McKinley Morganfield \\
    \textbf{\textcolor{darkgreen}{\fcolorbox{darkgreen}{white}{isSup}}}\ \textbf{Response}: \{'Reasoning': "The content states that McKinley Morganfield, known as Muddy Waters, is "often cited as the 'father of modern Chicago blues'," which specifically refers to his influence on the Chicago blues subgenre. However, the question asks who is considered the "father of the blues" in a general sense, which typically implies a foundational figure for the entire blues genre, not just a subgenre. Therefore, while Muddy Waters is a significant figure in blues, the content does not fully support the answer that he is the "father of the blues" overall; it only supports his title within the context of "modern Chicago blues." Thus, the answer "McKinley Morganfield" is not supported by the content for the broader question.", 'Response': 'False'\}
    
\end{quote}
\textbf{Reset content storage $\boldsymbol{C}$ and restart searching on next batch...}

\newpage
\section{Prompt template}
\label{app:prompt}
\subsection{isRel}
\begin{quote}
"""
You are an expert in reading comprehension and logical reasoning. Your task is to determine whether a given piece of text is related to a specific question.\\\\
Consider the following inputs:\\
1. Content:\ \{A piece of text provided to you.\}\\
2. Title:\ \{The title relevant to the text.\}\\
3. Question:\ \{A specific question that may or may not be related to the text.\}\\\\
Your response should include:\\
1. Reasoning: A step-by-step explanation of how you analyzed the text to determine if it is related to the question.\\
2. Response: A value (`True` or `False`) indicating whether the text is related to the question.\\\\
- **True**: The text is related and can help answer the question.\\
- **False**: The text is not related to the question.\\\\
Format your output as a JSON object:\\\\
-----\\
Schema\\
-----\\
\{\\
    "Reasoning": "Step-by-step reasoning explaining why the text is or isn't related to the question.",\\
    "Response": "True"/"False"\\
\}\\\\
-----\\
Example:\\
-----\\\\
1. Content: 'Photosynthesis is the process by which darkblue plants and some other organisms use sunlight to synthesize foods with the help of chlorophyll. The process converts carbon dioxide and water into glucose and oxygen.'\\
2. Title: 'Photosynthesis'\\
3. Question: 'What is the main product of cellular respiration?'\\\\
Output:\\
\{\\
    "Reasoning": "The question asks about the main product of cellular respiration, which is a process that converts glucose into energy, carbon dioxide, and water. The text provided discusses photosynthesis, a different process that converts carbon dioxide and water into glucose and oxygen. Since the text does not address cellular respiration, it is not related to the question.", "Response": "False"\\
\}\\\\
-----\\\\
1. Content: 'Cellular respiration is a set of metabolic reactions and processes that take place in the cells of organisms to convert biochemical energy from nutrients into adenosine triphosphate (ATP), and then release waste products.'\\
2. Title: 'Cellular Respiration'\\
3. Question: 'What is the main product of cellular respiration?'\\\\
Output:\\
\{\\
    "Reasoning": "The question asks about the main product of cellular respiration. The text directly discusses cellular respiration and mentions that it converts nutrients into ATP, which is the main product of the process. Therefore, the text is related to the question.", "Response": "True"\\
\}\\\\
-----\\
\\"""
\end{quote}
\subsection{QA prompt}
\begin{quote}
'''You are an expert in knowledge extraction, problem-solving, and reading comprehension. \\
You excel at identifying key information and providing concise answers.
\\\\
Given the following inputs:\\
1. Content: \{Relevant content that may contain the answer to the question.\}\\
2. Question: \{A specific question related to the content.\}
\\\\
Your task is to answer the question based on the provided content. The answer should be as simple as possible, typically an entity or a timestamp.
\\\\
Your response should only contain the answer itself. Do not explain, provide notes, or include any additional text, punctuation, or preposition (e.g., 'on', 'at'), or articles (e.g., 'a', 'an', 'the') unless absolutely necessary. 
\\\\
Output should be in the following JSON format:\\
\{\\
    "Reasoning": "A step-by-step solution trace explaining the reasoning behind the answer.",\\
    "Response": "The answer itself, as simple as possible."\\
\}
\\\\
-----\\
Example:\\
-----
\\\\
1. Content: 'The capital of France, known for its art, culture, and history, is Paris.'
2. Question: 'What is the capital of France?'
\\\\
Output:\\
\{\\
    "Reasoning": "The content directly states that the capital of France is Paris.",
    "Response": "Paris"\\
\}
\\\\
-----
\\\\
1. Content: 'The American Civil War ended on April 9, 1865, when General Robert E. Lee surrendered to General Ulysses S. Grant.'\\
2. Question: 'When did the American Civil War end?'
\\\\
Output:\\
\{\\
    "Reasoning": "The content provides a specific date for the end of the American Civil War.",
    "Response": "April 9 1865"\\
\}
\\\\
-----\\
\\"""
\end{quote}
\subsection{isSup}
\begin{quote}
"""You are an expert in critical analysis, problem-solving, and comprehension. You excel at evaluating the relationship between content and answers.
\\\\
Given the following inputs:\\
1. Content: \{A content specific to the question, describing the content of the image or text.\}\\
2. Question: \{A specific question.\}\\
3. Answer: \{The provided answer to the question.\}\\\\
Your task is to assess whether the content supports the given answer based on the question.\\\\
Please evaluate the support level of the content for the answer according to the following levels:\\
1. True: The answer is fully supported by the content.\\
2. Partial: The answer is partially supported by the content; it may be incomplete or not entirely accurate but contains some correct elements.\\
3. False: The answer is not supported by the content or is contradicted by it.\\\\
After evaluating, provide your verification and an explanation for your choice.\\\\
Output should be in the following JSON format:\\
\{\\
    "Reasoning": "A brief explanation of why the content fully, partially, or does not support the answer.",\\
    "Response": "True" / "Partial" / "False"\\
\}\\\\
Example:\\\\
-----\\\\
1. Content: "Tokyo is one of the most populous cities in the world and serves as the political, economic, and cultural center of Japan. The city is home to the Japanese government and the Imperial Palace."\\
2. Question: "What is the capital of Japan?"\\
3. Answer: "Tokyo."\\\\
Output:\\
\{\\
    "Reasoning": "The content clearly states that Tokyo is the political center of Japan and houses the government, fully supporting the answer that Tokyo is the capital of Japan.",
    "Response": "True"\\
\}\\\\
-----\\\\
1. Content: "World War II, which began in 1939, was a global conflict that involved most of the world's nations. The war saw the rise and fall of major powers and resulted in significant geopolitical changes. The conflict formally ended with the unconditional surrender of Germany in May 1945, followed by the surrender of Japan in September 1945."\\
2. Question: "What year did World War II end?"\\
3. Answer: "1944."\\\\
Output:\\
\{\\
    "Reasoning": "The content provides information that World War II ended in 1945, contradicting the given answer of 1944. Therefore, the content does not support the answer.",
    "Response": "False"\\
\}\\\\
-----\\\\
1. Content: "The Amazon Rainforest, covering approximately 5.5 million square kilometers, is often referred to as the 'lungs of the Earth.' It plays a critical role in regulating the global climate by absorbing large amounts of carbon dioxide. The rainforest is home to an incredible diversity of flora and fauna, many of which are endemic to the region."\\
2. Question: "How much area does the Amazon Rainforest cover?"\\
3. Answer: "The Amazon Rainforest plays a crucial role in regulating the global climate."\\\\
Output:\\
\{\\
    "Reasoning": "The answer correctly mentions the role of the Amazon Rainforest in climate regulation, but it does not address the specific question about the area it covers. Therefore, the answer is only partially supported by the content.",\\
    "Response": "Partial"\\
\}\\
-----
\\\\"""
\end{quote}

\subsection{isUse}
\begin{quote}
"""You are an expert in critical analysis, problem-solving, and comprehension. \\
Your task is to evaluate whether a given answer appropriately addresses a specific question based on the provided content.\\\\
Given the following inputs:\\
1. Content: \{The source material from which the answer is derived.\}\\
2. Question: \{A specific question that needs to be answered.\}\\
3. Answer: \{The provided answer in response to the question.\}\\\\
Your goal is to determine whether the answer is appropriate by considering the following criteria:\\\\
1. **Format Accuracy**: The answer should be in the expected format. For example, if the question asks for a date, the answer should be presented in a date format.\\
2. **Logical Consistency**: The answer must be logically correct and align with the information provided in the content.\\\\
After your evaluation, provide a detailed explanation of your reasoning, breaking it down step by step, and offer a clear conclusion.\\\\
Your output should be in the following JSON format:\\
\{\\
    "Reasoning": "A comprehensive explanation of whether the answer meets the expected format and logical criteria, supported by specific references to the content.",\\
    "Response": "True" / "False"\\
\}\\\\
-----\\
Examples:\\
-----\\\\
1. Content: 'Mount Everest, located in the Himalayas on the border between Nepal and the Tibet Autonomous Region of China, is the highest mountain in the world, with a peak that reaches 8,848 meters (29,029 feet) above sea level.'\\
2. Question: What is the height of the highest mountain in the world?\\
3. Answer: 8,848 meters\\\\
Output:\\
\{\\
    "Reasoning": "The answer correctly identifies the height of Mount Everest as 8,848 meters, which is both in the expected numerical format and logically consistent with the provided content.",\\
    "Response": "True"\\
\}\\\\
-----\\\\
1. Content: 'The Nile River, which flows through northeastern Africa, is considered one of the longest rivers in the world, with a length of approximately 6,650 kilometers (4,130 miles). The river is a major waterway for countries such as Egypt and Sudan.'\\
2. Question: How many countries does the Nile River flow through?\\
3. Answer: The Nile River is 6,650 kilometers long.\\\\
Output:\\
\{\\
    "Reasoning": "The answer mentions the length of the Nile River but fails to address the question, which asks for the number of countries the river flows through. The answer is neither in the expected format nor logically consistent with the question.",\\
    "Response": "False"\\
\}\\\\
------
\\\\"""
\end{quote}

\subsection{Image inference prompt}
\begin{quote}
"""You are an expert at problem solving, knowledge extraction, and reading comprehension.\\
You excel at identifying requirements, breaking tasks down, and solving problems step-by-step.\\\\
Given the following inputs:\\
1. Title: \{Title of the context\}\\
2. Question: \{A given question\}\\\\
Your objective is to determine whether the image content is directly related to answering the question. If the image and title are relevant, you should provide a thorough, step-by-step reasoning that clearly demonstrates the connection between the image content and the question. If the image is not relevant, explain why there is no connection.\\\\
Please follow these steps:\\
1. **Analyze the Title and Image**: Assess the image content based on the title and determine its relevance to the question.\\
2. **Develop Reasoning**: If the image is relevant, provide a detailed and logical explanation of how the image content answers the question. If irrelevant, explain the lack of connection between the image and the question.\\
3. **Conclude with a Response**: Clearly state whether the image is relevant (`True`) or irrelevant (`False`) to the question.\\\\
Your output should adhere to the following JSON schema:\\\\
-----\\
Schema\\
-----\\\\
\{\\
    "Reasoning": "A detailed and logical step-by-step explanation of why the image content is or is not related to the question.",\\
    "Response": "True" / "False"\\
\}\\\\
-----\\
Examples\\
-----\\\\
1. Title: 'The Cosby Show'\\
2. Question: 'What color is the inside of the text in The Cosby Show?'\\\\
Output:\\
\{\\
  "Reasoning": "The image titled 'The Cosby Show' prominently displays the show's title. The inside of the text is white, standing out against a darker background. This contrast enhances visibility and aligns with the show's branding, making the title easily recognizable. The use of white inside the text reflects the cultural significance of the show's branding.",\\
  "Response": "True"\\
\}\\\\
-----\\\\
1. Title: 'Mercy (TV series)'\\
2. Question: 'Is the background to the Person of Interest (TV series) poster colored or colorless?'\\\\
Output:\\
\{\\
  "Reasoning": "The image titled 'Mercy (TV series)' features a grayscale background, depicting a cityscape or a map-like layout. The only color in the image is a red triangular symbol, with the rest of the design focusing on muted tones. The colorless background reinforces the show's themes of surveillance and intrigue, creating a serious and dramatic tone. Therefore, the background is predominantly colorless. However, the question asking about the Person of Interest (TV series) poster, not about the Mercy (TV series), so the image does not seem to be related to the question.",\\
  "Response": "False"\\
\}\\\\
-----
\\\\"""
\end{quote}

\end{document}